\documentclass[letterpaper, 10 pt, conference]{ieeeconf}  

\IEEEoverridecommandlockouts                              

\usepackage{graphics} 
\usepackage{epsfig} 

\title{\LARGE \bf
Enabling Depth-driven Visual Attention on the iCub Humanoid Robot: Instructions for Use and New Perspectives
}

\author{Giulia Pasquale$^{1,2,3}$, Tanis Mar$^{1,3}$, Carlo Ciliberto$^{2,4}$, Lorenzo Rosasco$^{2,3,4}$, Lorenzo Natale$^{1}$
\thanks{$^{1}$iCub Facility, Istituto Italiano di Tecnologia}%
\thanks{$^{2}$Laboratory for Computational and Statistical Learning, Istituto Italiano di Tecnologia}
\thanks{$^{3}$ Dipartimento di Informatica, Bioingegneria, Robotica e Ingegneria dei Sistemi, Universit\'a degli Studi di Genova}%
\thanks{$^{4}$ Poggio Lab, Massachusetts Institute of Technology}%
}

\begin{document}

\maketitle
\thispagestyle{empty}
\pagestyle{empty}

\begin{abstract}

The importance of depth perception in the interactions that humans have within their nearby space is a well established fact. Consequently, it is also well known that the possibility of exploiting good stereo information would ease and, in many cases, enable, a large variety of attentional and interactive behaviors on humanoid robotic platforms. However, the difficulty of computing real-time and robust binocular disparity maps from moving stereo cameras often prevents from relying on this kind of cue to visually guide robots' attention and actions in real-world scenarios. The contribution of this paper is two-fold: first, we show that the Efficient Large-scale Stereo (ELAS) Matching algorithm~\cite{LIBELAS} for computation of the disparity map is well suited to be used on a humanoid robotic platform as the iCub robot; second, we show how, provided with a fast and reliable stereo system, implementing relatively challenging visual behaviors in natural settings can require much less effort. As a case of study we consider the common situation where the robot is asked to focus the attention on one object close in the scene, showing how a simple but effective disparity-based segmentation solves the problem in this case. Indeed this example paves the way to a variety of other similar applications.

\end{abstract}

\section{Introduction}
\label{sec:intro}

Depth perception is a fundamental ability for physical agents operating in unstructured environments. Indeed, even basic tasks such as reaching for an object or navigation require a sufficiently accurate model of the three-dimensional structure of the scene in order to be carried out efficiently. 

The problem of $3$D reconstruction is particularly relevant in robotics. Indeed, in this setting perception often represents the main bottleneck for most applications that require interaction with the environment. In humanoid robotics, a typical solution is to employ stereo vision to match the scene observed from the two cameras mounted on the robot's ``eyes'' and then model the $3$D structure of the scene by means of triangulation.

Typically, the main obstacle to stereo vision lies in the process of matching the $2D$ points from one image to the other in order to compute the amount of displacement, or {\it disparity}. In this work we consider the Efficient Large-scale Stereo (ELAS) Matching algorithm~\cite{LIBELAS}, and incorporate it in the visual perceptual system of the iCub robot~\cite{icub}. 

According to standard {\it KITTI Stereo-Vision Benchmark}~\cite{KITTI}, ELAS offers a reasonable trade-off between quality of the disparity estimation ($54^{th}$ out of $78$) and computational times ($13^{th}$ out of $78$), which makes it particularly suited for applications that require real-time performance. Moreover, the algorithm available as the library {\it LIBELAS}\footnote{www.libelas.com} is open, self-contained and already highly optimized without need for specific accelerators.

Our main contributions are: $1)$ to have incorporated and tested the LIBELAS library on the iCub robot making it readily available for the iCub community, which we believe could greatly benefit from this algorithm; $2)$ to present a set of quantitative and qualitative experiments to assess the efficacy of the ELAS algorithm in a realistic robotics setting. 

\section{Related work}
\label{sec:stateart}

Depth is a natural cue to be used when the robot's attention needs to be focused on close entities in its workspace. For example, consider a very common situation for a humanoid robotic platform, like the one where a human stands in front of the robot showing to it an object to be recognized or grasped. Both motion- and appearance-based approaches to focus the robot's attention on the object of interest would impose many constraints on this even simple Human-Robot Interaction (HRI) scenario.
Indeed, color-based methods work under strict assumptions on the light conditions, kind of background (preferably a table or a wall) and generally fail in cluttered settings. Model-based methods, beyond being affected to some extent by the same limitations, need a model of the object to be known a-priori. Motion-based methods (see, e.g.,~\cite{motionCUT}) work under the obvious assumption that the objects are moving, the speed of the object being often critical for the detection. Instead, when the robot is required to look at something we are showing to it, or which is located nearby, the most distinguishing feature is simply the fact that the object of interest is closer to the robot than the background.

For this reason, depth information has been exploited in a variety of robotics applications in the past~\cite{goerick05,wersing06,wersing07,wersing08,goerick06,rudinac12}. However, it is not easy to find methods for depth estimation from a stereo pair which are a good trade-off between robustness (e.g. to lighting conditions) and speed, two requirements that are key for working in real-world robotic scenarios. Therefore, alternative solutions as for example Kinect RGB-D sensors have been adopted, even when working on the iCub humanoid, which is equipped with a human-like stereo camera system (see \cite{lyubova2015passive}). The main motivation of this work is thus to ``upgrade'' the iCub robot's depth perception and to show that this improvement opens the way to a range of possible applications where disparity and depth can be successfully used to visually guide the robot's attention and actions also without relying on a Kinect.

We decided to rely on the LIBELAS library because, comparing to other local dense stereo matching methods, which can be faster (see e.g. OpenCV's Block Matching algorithm implementation~\cite{OpenCV} for which a GPU accelerated version is also available), LIBELAS has been shown to provide better matching results in texture-less regions and to be more robust against illumination changes (see the KITTI Stereo-Vision Benchmark~\cite{KITTI}), which are common factors in robotic settings, while still being competitive in the computational time. Indeed, according also to recent experimental evidence~\cite{SinhaCVPR2014}, when compared to semi-global methods, including OpenCV's implementation of Hirschmuller's Semi-Global Block Matching (SGMB) algorithm~\cite{SGBM}, currently in use on the iCub platform, LIBELAS scales much better with respect to the image resolution and the disparity range.

These are also the main reasons why LIBELAS has been the library of choice for many previous robotic applications. See for example ~\cite{Tombari2011}, where this algorithm is evaluated with respect to an object recognition task, or~\cite{Bergh2012}, where LIBELAS is used in conjunction with color and optic flow to provide a real-time super-pixel segmentation of the scene, or~\cite{mitzel2012close} and~\cite{baumgartner2013tracking}, where LIBELAS provides the depth map which is used to detect people or also infer the interactions between people and objects. 

Moreover, in~\cite{stereo4impaired_2012} it is shown that this algorithm is suited to be implemented on an embedded ARM-based processor and still runs in real-time on mid-resolution images. This is particularly important in the perspective of having wireless communication between the robot body and the external computational nodes: indeed, in order to reduce the data to be transmitted, the first step would be to move low-level visual computation on the processor mounted on the robot's head. Therefore we tested LIBELAS in indoor settings on the iCub and integrated it within the iCub stereo vision repository (YARP Robotology - Stereo Vision~\cite{robotology_stereovision}) publicly available for the iCub community.

In Sec.~\ref{sec:methods1} we briefly review the processing steps currently adopted on the iCub for estimating the scene depth from the moving stereo pair given by its eyes; in Sec.~\ref{sec:methods2} we describe a simple application that we devised to focus the robot's attention on the closest object in its workspace. Finally, the experimental results that we report on the iCub robot in Sec.~\ref{sec:results} demonstrate the effectiveness of the proposed approach, paving the way further applications exploiting the improvement in the disparity computation.

\section{Depth Estimation}
\label{sec:methods1}

In this section we briefly describe the depth estimation pipeline adopted in this work. Following the standard approach from multi-view geometry~\cite{hartley03}, this process is organized into two main phases: image rectification and disparity computation.

The rectification step estimates the geometrical transformation matrix relating left and right image planes in order to align the epipolar lines with the image scanlines. After this operation, the (horizontal) disparity computation can be carried out for each pixel in the left (right) rectified image, by searching its correspondent point in the right (left) rectified image along its scanline. The resulting disparity map provides an estimation of the $3$D structure of the scene as a cloud of points (whose projections end up on the image pixels) with respect to the observer. To recover the $3$D position of the point corresponding to a specific pixel, the camera's extrinsic parameters can be used, in combination with its disparity, to re-project it.

Regarding the estimation of the camera parameters, we follow the procedure described in~\cite{fanello14depth}, which allows both to pre-compute them during an initial calibration phase and then to re-calibrate them at runtime if needed (for instance when the iCub's eyes are moved and the relative pose of the left and right cameras changes). As mentioned already, for disparity estimation we adopt the Efficient Large-scale Stereo (ELAS) Matching algorithm proposed in~\cite{LIBELAS}.

\subsection{Rectification}
\label{sec:rectification}

Image rectification consists in the process of transforming a set of multiple images onto the same plane and is a fundamental step to most depth estimation algorithms. Rectification requires knowledge of both the intrinsic (camera specific) parameters of the two (or more) cameras and extrinsic parameters, i.e. the position and orientation of the cameras with respect to the world reference frame. More formally, any $3$D point with coordinates $\mathbf{X} = (x,y,z,1)^\top$ with respect to the world reference frame, is mapped on the camera image plane $\mathbf{x} = (u,v,1)^\top$ via the transformation 
\begin{equation}
\label{eq:projection}
s\mathbf{x} = P\mathbf{X}
\end{equation}
where $s\in\mathbf{R}$ is a scaling factor and the {\em Projection Matrix} $P\in\mathbf{R}^{3\times4}$ can be factorized as $P = K[R|t]$ with $K\in\mathbf{R}^{3\times3}$ and $[R|t]\in\mathbf{R}^{3\times4}$ respectively the matrices of intrinsic and extrinsic parameters. 

Both sets of camera parameters can in general be estimated offline during a calibration phase. However, while intrinsic parameters are camera specific and do not change over time, on the iCub the relative pose of the cameras changes whenever the robot's eyes vergence or pan is modified. Moreover, due to elasticities and backlash, the relative pose of the two cameras can slightly change whenever the robot moves (even if the eyes are kept fixed). To circumvent this issue, in~\cite{fanello14depth} the authors precomputed the intrinsic parameters matrices $K_L$ and $K_R$ (via standard calibration procedure as in~\cite{hartley03}), while performing the extrinsic parameters calibration at runtime. Such a calibration was carried out by employing a SIFT matching algorithm to estimate the {\em Fundamental Matrix} between the two camera planes. We refer the reader to~\cite{fanello14depth} for more details. However, in order to achieve real-time performance, the authors exploited the the known robot's kinematics to approximate the camera transformation between subsequent frames and perform the re-calibration via SIFT matching at a lower frame rate. This procedure is implemented in the {\em SFM (structure from motion)} module included in the iCub stereo vision repository~\cite{robotology_stereovision}).

Once the projection matrices $P_L$ and $P_R$ associated to the left and right cameras are known, the corresponding images can be mapped onto the same plane, i.e. they are {\em rectified}. They are therefore ready for the subsequent stage: disparity estimation.

\subsection{Disparity Computation with ELAS}
\label{sec:disparity_computation}

Disparity estimation consists in the process of evaluating the displacement of pixels from one (rectified) image to the other. Disparity is usually computed after rectification since at this stage the corresponding image points from the left and right camera lie on the same scanline and therefore matching can be restricted to those horizontal lines. A variety of disparity estimation methods have been proposed in the literature. In this paper we have adopted the Efficient Large-Scale Stereo (ELAS) Matching algorithm proposed in~\cite{LIBELAS}, which consists in the following two phases:
\begin{enumerate}
\item A set of robust support points is detected and matched across the two images.
\item The support points are used within a Bayesian framework to determine the most likely disparity values of all points on a predefined grid set on the image plane.
\end{enumerate}
In this section we offer a very brief overview of the ideas underlying ELAS while referring the reader to the original paper for a more detailed description of the algorithm~\cite{LIBELAS}.

\subsubsection{Support Points}
The first phase of ELAS is performed on a predetermined grid on the image plane, where candidate points are selected depending on their local appearance. To do so, the authors used a vector of local orientations (response to oriented Sobel filters) and performed robust matching between such feature vectors to eliminate unstable pairs of points. The outcome of this stage is a set $\mathbf{S}$ of points $s = (u,v,d)^\top$ which encode the position $(u,v)$ of a support point on the left (rectified) image and the the disparity $d$ with respect to the matched point on the right (rectified) image.

\subsubsection{Bayesian Inference}
The second phase relies on the two-view geometry parameters estimated in Sec.~\ref{sec:rectification} and the support points $\mathbf{S}$ to predict the most likely disparity values for the remaining image pixels. In particular, the authors adopt a Bayesian framework to model the likelihood 
$$
p(d | x^{(L)}, x_1^{(R)}, \dots, x_n^{(R)}, \mathbf{S})
$$
to observe a disparity $d$ for a given point $x^{(L)}$ on the left image and a set of candidate corresponding points $x_1^{(R)}, \dots, x_n^{(R)}$ on the right image. The most likely disparity value is therefore estimated by factorizing such likelihood and performing a Maximum a-posteriori (MAP) procedure.

As pointed out by the authors in~\cite{LIBELAS}, this procedure can be carried out independently for each image point and indeed it is fast and parallelizable. Clearly, this is a critical feature for the robotic setting, where the disparity estimation process must be computed at real-time, possibly at frame rate. Therefore we opted for LIBELAS library and in particular for its OpenMP parallelization, available at the same website.

\section{Depth-driven Visual Attention System}
\label{sec:methods2}

In order to asses the efficacy of the disparity map provided by ELAS in a typical humanoid robotics setting, here we consider a benchmark application. We designed a segmentation procedure based on the disparity map produced by the pipeline reviewed in Sec.~\ref{sec:methods1} to identify distinct three-dimensional entities in the scene and focus the robot's gaze toward the object that lied closest to the camera stereo pair (i.e. the iCub's eyes). By following this strategy we were able to implement a simple but effective tracking algorithm that would continuously focus the robot's attention and gaze towards the closest object in the scene, while at the same time providing also an approximate visual segmentation. 

We first employed this basic tracking system to perform a qualitative and quantitative analysis of the disparity map produced by ELAS in a real-world indoor robotics setting. 
Then, within this general scenario we defined a reliable protocol to acquire ground-truth for visual object recognition. Indeed, this approach can be employed to acquire a dataset of images depicting multiple objects held in the hand of a human teacher while he/she shows them to the iCub. A similar strategy, based on independent motion detection rather and disparity-driven attention, was indeed previously employed on the iCub robot~\cite{fanello13c,fanello13b}. To this regard, we will show that in such an application disparity information results in a more reliable and stable cue.

In the following we describe the algorithm we devised to segment the object closest to the iCub cameras and the two applications of such information for tracking and ground-truth acquisition.

\subsection{Foremost Object Segmentation}
\label{sec:segmentation}

In order to cope with the real-time requirement imposed by the robotic setting, given that disparity computation per se is computationally onerous, we had to reduce the post-processing operations on the disparity map at minimum. Therefore, the devised segmentation algorithm is the simplest one that could provide us with what was needed for this kind of application, i.e., a reasonably stable and accurate blob around the closest proto-object in the scene. Nonetheless, we are aware of the existence of more sophisticated algorithms, which may provide more precise segmentations (see, e.g.~\cite{disp_segmentation_Li2012}), that could be easily plugged in the present pipeline to realize different possible of behaviors.

The basic steps of the algorithm are:

\begin{itemize}
\item{\textit{Filtering} A $5 \times 5$ Gaussian filter ($\sigma_{x}=\sigma_{y}=1.5$) is applied to the disparity map before thresholding it to suppress the pixels under a certain value (set to $50$). Then follows a sequence of $4$ dilation and $2$ erosion operations, interleaved by another $5 \times 5$ Gaussan filtering ($\sigma_{x}=\sigma_{y}=2$), to suppress noisy smallest blobs and fill holes in the bigger ones. }
\item{\textit{Blob selection} A simple routine, iteratively:
\begin{enumerate}
\item{finds the location of the brightest pixel}
\item{grows its surrounding pixels until their value is comprised between $ v-\frac{v}{u_{-}} $ and $ v+\frac{v}{u_{+}} $, being $v$ the value of the seed pixel and $u_{+}=20$ and $u_{-}=10$ two relative thresholds}
\item{suppresses (putting to zero) the found region if its size is lower than a threshold (set to an approximate amount of $20 \times 20 $ pixels for $320 \times 240$ images and of $40 \times 40 $ pixels for $640 \times 480$ images)}
\item{starts again until a blob satisfying the size requirement is found}
\end{enumerate}}
\item{\textit{Computing the blob's centroid and ROI} After double-checking that a single and big enough blob is selected (and not, e.g., two connected blobs), its center of mass and the smallest rectangular enclosing bounding box (with an arbitrary margin) are computed.}
\item{\textit{Averaging over a temporal buffer} Finally, the centroid and the ROI are averaged over a buffer of $n$ frames (with e.g. $n=3$) in order to avoid isolated mis-detections.}
\end{itemize}

At this point the detected centroid is triangulated and passed to the gaze controller in order to make the robot focus on it. This pipeline is looped in real-time so that the robot is able to follow the closest object with the gaze.

It is then clear that relying on a fast and robust disparity map (eventually at the expenses of some sub-pixel precision) in this kind of application is critical, and the reported results confirm that LIBELAS is suited to this task.

We note also that this is a very basic (yet effective) implementation for a disparity-driven attention system and that further improvements e.g., applying a Kalman filter to the trajectory of the $3$D centroid, could be introduced to smooth and stabilize the resulting tracking system.

\section{Experimental evaluation}
\label{sec:results}

In this section we present a qualitative as well as quantitative analysis of the depth estimation process described in Sec.~\ref{sec:methods1}, with particular focus on the improvements provided by the ELAS algorithm, which represents the novel element of the pipeline for disparity computation. As we are mainly concerned in assessing the possibility to employ this algorithm in real-time, real-world robotics applications, we first evaluate the disparity-based segmentation protocol introduced in Sec.~\ref{sec:segmentation} and then we study the efficacy of such an approach towards a depth-driven visual attention behavior. 

For our experiments, we employed the OpenCV~\cite{OpenCV} implementation of the Semi-Global Block Matching algorithm (SGBM)~\cite{SGBM} as a baseline to compare the performance of ELAS. Indeed, SGBM was considered the ``off-the-shelf'' disparity estimation algorithm for the iCub robot used in the {\it SFM} module in the iCub stereo vision repository (see \cite{fanello14depth}). Our analysis in the following suggests that SGBM should be actually replaced by ELAS. Indeed we have recently updated the SFM module to allow for both algorithms to be used (default is now ELAS).

\subsection{Real-time depth segmentation}

\begin{figure*}[thpb]
\centering
\includegraphics[scale=0.4]{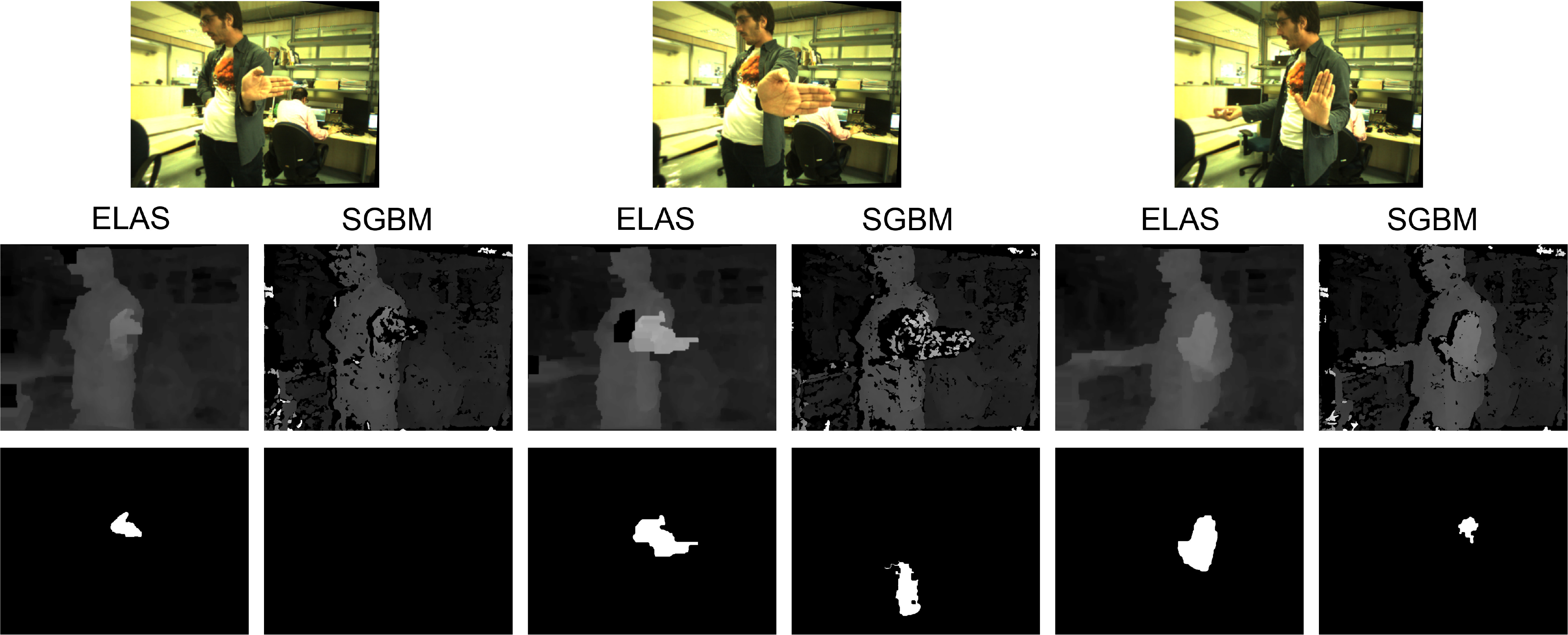}
\caption{Examples from a sequence recorded on the iCub's cameras, with fixed eyes and head. Top: left rectified images. Middle: disparity maps computed by LIBELAS and OpenCV's SGBM methods. Bottom: segmentation of the closest blob in the scene.}
\label{fig:carlos_hand}
\end{figure*}

In Fig.~\ref{fig:carlos_hand} we report three examples of images (at $640 \times 480$ resolution) sampled from $200$ frames acquired across $2$ minutes and recorded from the iCub cameras while a human subject was moving his hand in front of the robot. We report the performance of the disparity-based segmentation protocol described in Sec.~\ref{sec:segmentation}: the first row of Fig.~\ref{fig:carlos_hand} depicts the rectified images acquired from the left camera (those from the right camera are not reported); the second and third rows report respectively the disparity maps and the resulting segmentation, obtained with ELAS (odd columns) or SGBM (even columns).

From Fig.~\ref{fig:carlos_hand} it can be noticed that ELAS is in general more robust and computationally more efficient, allowing to successfully detect the hand of the operator while SGBM often fails (see, e.g., the first and second frames). In particular, regarding computational efficiency, in Table~\ref{tab:carlo_times} we report the computational time required on our platform (Intel(R) Core(TM) i$7$ $3770$QM CPU at $3.40$GHz with $16$GB RAM) to perform the disparity estimation and segmentation, averaged over the whole acquisition sequence. We also report the ratio of ``missed'' blobs: the ratio of frames for which the segmentation algorithm failed to detect any blob (as it happens with SGBM in the first frame of Fig.~\ref{fig:carlos_hand} or returned a wrong one (as it happens, again with SGBM, in the second frame of Fig.~\ref{fig:carlos_hand}).

It can be noticed that the LIBELAS implementation is fast (achieving a $15$ fps rate with respect to the $5$ fps provided by SGBM) and robust enough to allow for further applications of the computed disparity, such as the depth-driven attention behavior described in the following. 

\begin{table}[h]
\caption{Average computational times and percentage of blobs missed by LIBELAS and SGBM over the sequence represented in Fig.~\ref{fig:carlos_hand}.}
\label{tab:carlo_times}
\begin{center}
\begin{tabular}{c|cc}
& SGBM & ELAS\\
\hline
Time Disp [ms] & 190 & 60\\
Time Segm [ms] & 20 & 5\\
Time Tot [ms] & 210 & 65\\
Missed Blobs [\%] & 11.2 & 2\\
\end{tabular}
\end{center}
\end{table}

We conclude our qualitative analysis by reporting the specific parameters chosen for our experiments: the disparity range was set to $[0,127]$ for both ELAS and SGBM. In Table~\ref{tab:parametersSGBM} we report the parameters of the SGBM algorithm which have been tuned to the specific iCub's indoor setting. Those not reported were left to their default value (see OpenCV's documentation~\cite{OpenCV}).
\begin{table}[h]
\caption{SGBM parameter setting.}
\label{tab:parametersSGBM}
\begin{center}
\begin{tabular}{l|c}
Parameter & Value \\
\hline
preFilterCap & $63$ \\
SADWindowSize & $7$ \\
P1 & $8*7*7$ \\
P2 & $32*7*7$ \\
uniquenessRatio & $15$ \\
speckleWindowSize & $50$ \\
speckleRange & $16$ \\
disp12MaxDiff & $0$ \\
\end{tabular}
\end{center}
\end{table}
In the case of the LIBELAS implementation, we chose the \textit{$MIDDLEBURY$} preset of parameters offered by the library. In order to speedup computations we set the \textit{$post\_process\_only\_left$} and the \textit{$subsampling$} parameters to true and employed the {\it OpenMP} accelerated version of the library. All other parameters were left to default values.

\subsection{Depth-driven visual attention}
\label{sec:tracking}

In this section we consider a simple yet effective attention system driven by disparity information, whose underlying principle is to keep the robots's gaze focused on the closest object in the scene. In particular we consider the following setting: a human actor standing in front of the robot exhibits an object in front of the robot cameras and then moves it, without a pre-fixed trajectory, in order to evaluate the stability of the resulting ``tracking'' application.

We employed the pipeline described in Sec.~\ref{sec:methods2}. Therefore, the centroid of the blob on the left image plane, obtained by the disparity segmentation module, was re-projected to its corresponding $3$D position in the Cartesian space. Finally, the $3$D point was fed to the module in charge of controlling the robot's gaze ({\it iKinGazeCtrl~\cite{iKinGazeCtrl}}), which moved the robot's eyes in order to fixate the specified $3$D point. As a consequence, the head and the eyes of the robot position were continuously updated to keep the focus of attention fixed on the required target, i.e. the closest object in the visual field, while the human actor was moving it in front of the iCub cameras.

In the current experiment we used low-resolution images (with resolution $320\times240$) and the disparity range was reduced to $[0,95]$. We used the same parameters for SGBM, whereas for LIBELAS there was no need to enable the \textit{$subsampling$} since the lower resolution already allowed to achieve frame-rate performance ($30$ fps) on our platform. Experiments with SGBM were performed offline since the lower efficiency ($\sim 10$ fps) and did not allow for a smooth tracking. 

We compared the result of the disparity-based segmentation with the output of a model-based object tracker~\cite{taiana2010tracking}. We used a red ball (see Fig.~\ref{fig:wrong1_disp}) for which a particle filter tracker is already implemented in the iCub repository and which uses color and $3$D shape features. As the operator moved the red ball in front of the robot, the gaze was focused towards it (since it was the closest object in the scene) and information about its estimated position was acquired independently using the disparity-based segmentation procedure described above and the colour/shape-based particle filter tracker. More accurately, we recorded the coordinates $(u_{disp},v_{disp})$ of the closest blob's centroid on the left image plane, provided at each frame by the segmentation module on top of ELAS disparity map, and the coordinates $(u_{model},v_{model})$ of the center of the red ball in the same image plane, provided by the red ball detector.

\begin{figure}[t]
\centering
\includegraphics[width=1.0\linewidth]{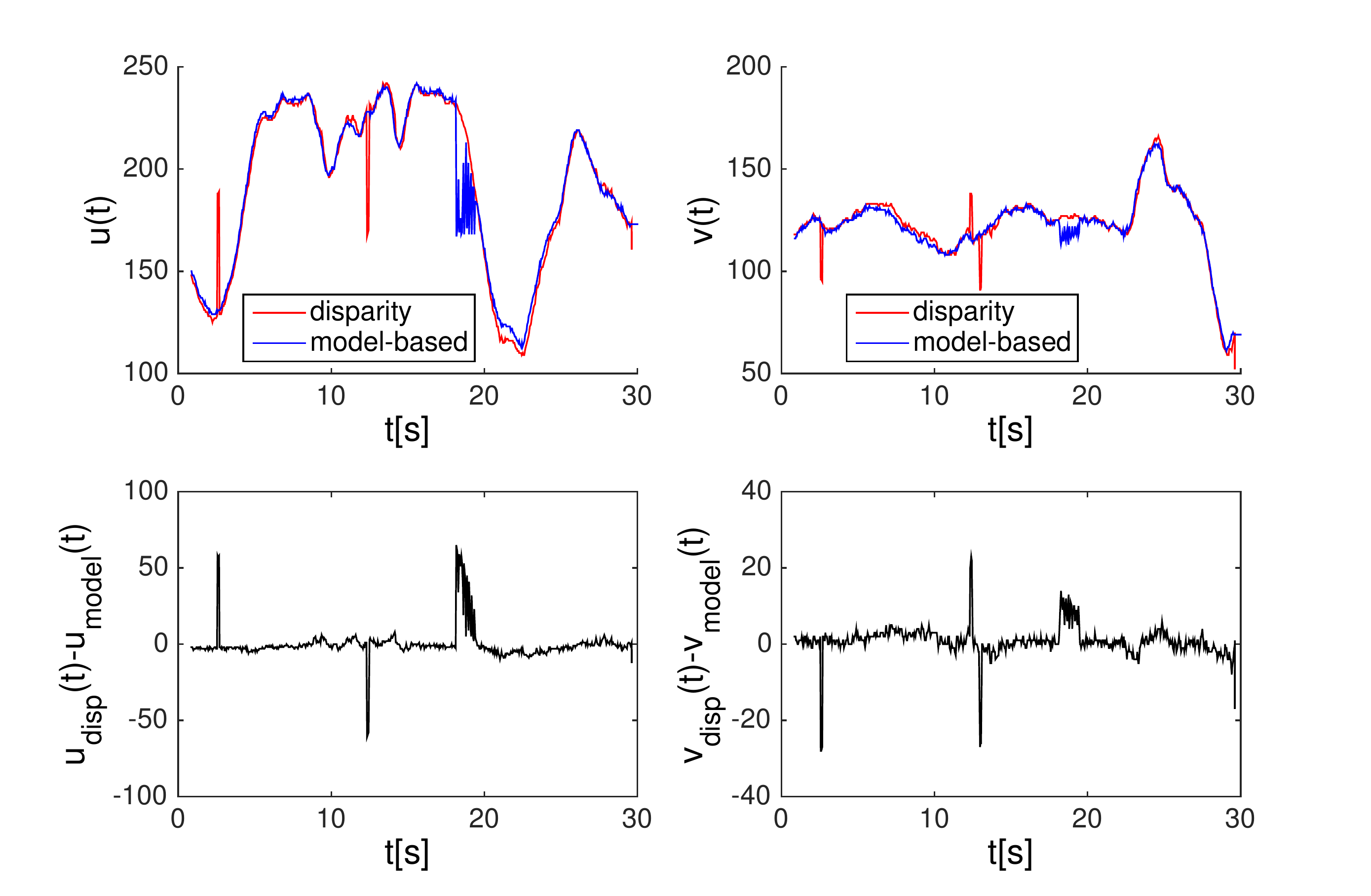}
\caption{Top: coordinates of the closest blob's centroid on the left image plane (red trace), provided at each frame by the disparity segmentation module, and of the center of the red ball in the same image plane (blue trace), provided by the red ball detector. Bottom: difference between the two. LIBELAS is used to provide the disparity map.}
\label{fig:dataVSblobs_acquired}
\end{figure}

\begin{figure}[t]
\centering
\includegraphics[width=1.0\linewidth]{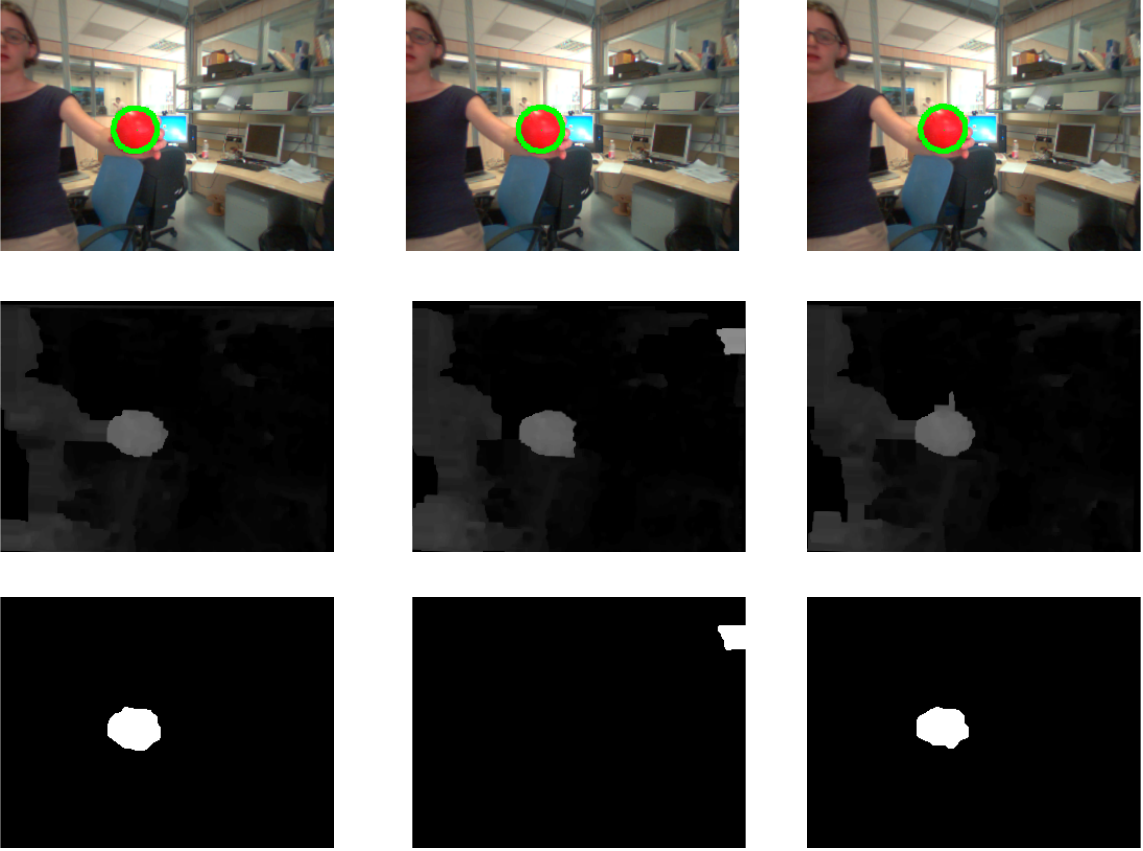}
\caption{Frames extracted from the sequence represented in Fig.~\ref{fig:dataVSblobs_acquired} around $t=2.6s $, when LIBELAS fails to detect the closest object. Top: output of the read-ball tracker. Middle: disparity map. Bottom: disparity segmentation. }
\label{fig:wrong1_disp}
\end{figure}

\begin{figure}[t]
\centering
\includegraphics[width=1.0\linewidth]{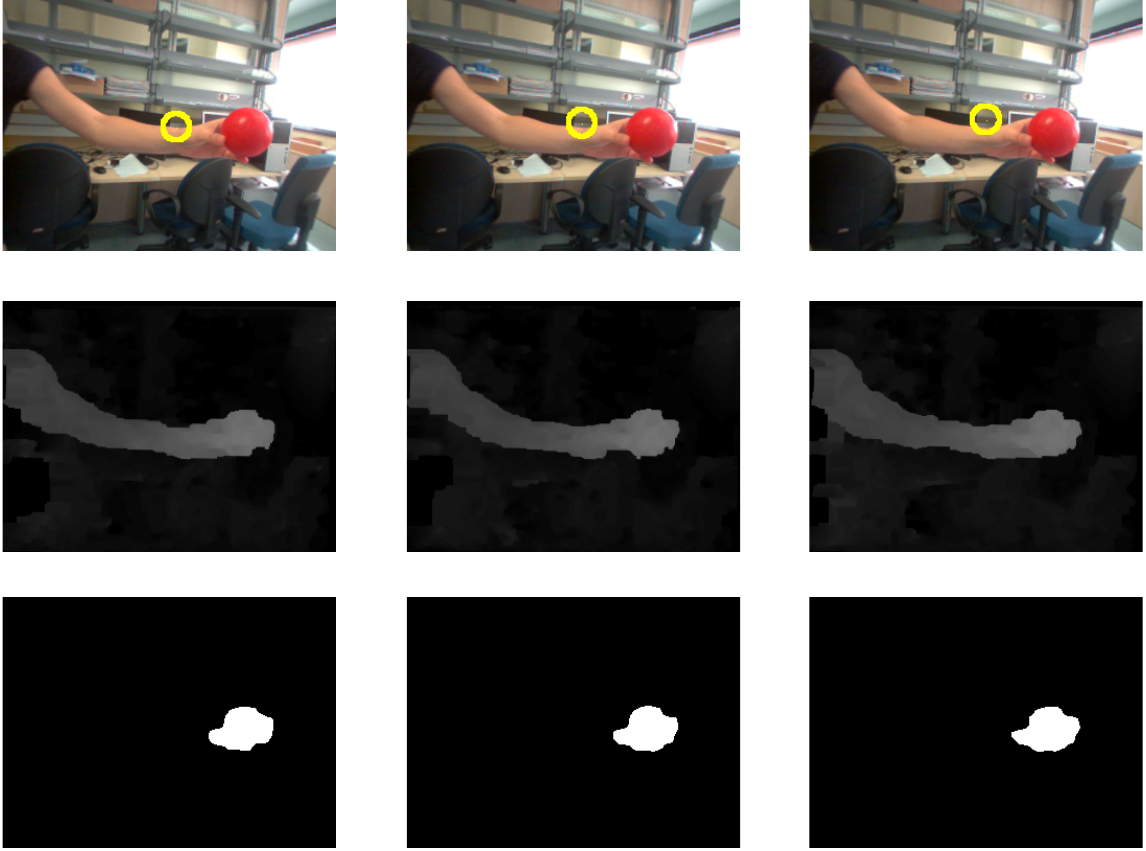}
\caption{Frames extracted from the sequence represented in Fig.~\ref{fig:dataVSblobs_acquired} in the period from $t=18s$ to $t=20s$, when the red-ball tracker fails to detect its target. Top: output of the tracker. Middle: disparity map. Bottom: disparity segmentation.}
\label{fig:wrong_pf}
\end{figure}

Fig.~\ref{fig:dataVSblobs_acquired} reports the image plane coordinates (top rows) with red and blue colors respectively for disparity and model-based tracker and their difference (bottom rows). Notice that for some frames ELAS estimates a wrong position for the blob (sudden jumps in the red curves); similarly, the red-ball tracker fails to detect its target within a $2s$ interval around $t=18s$ in the plot. In Fig.~\ref{fig:wrong1_disp} we provide a short sequence showing the ELAS failure around $t=2.6s$. Notice that the error affects an isolated frame and can be removed by simply filtering the $2$D image position detected by raw segmentation. In Fig.~\ref{fig:wrong_pf} we report instead a short sequence extracted from the interval in which the red-ball detector fails: in this case the error is due to a constant mis-detection caused by the slightly adverse lighting conditions. This shows how the disparity cue for tracking and segmentation can be in general more robust than appearance-based information.

\begin{figure}[t]
\centering
\includegraphics[width=1.0\linewidth]{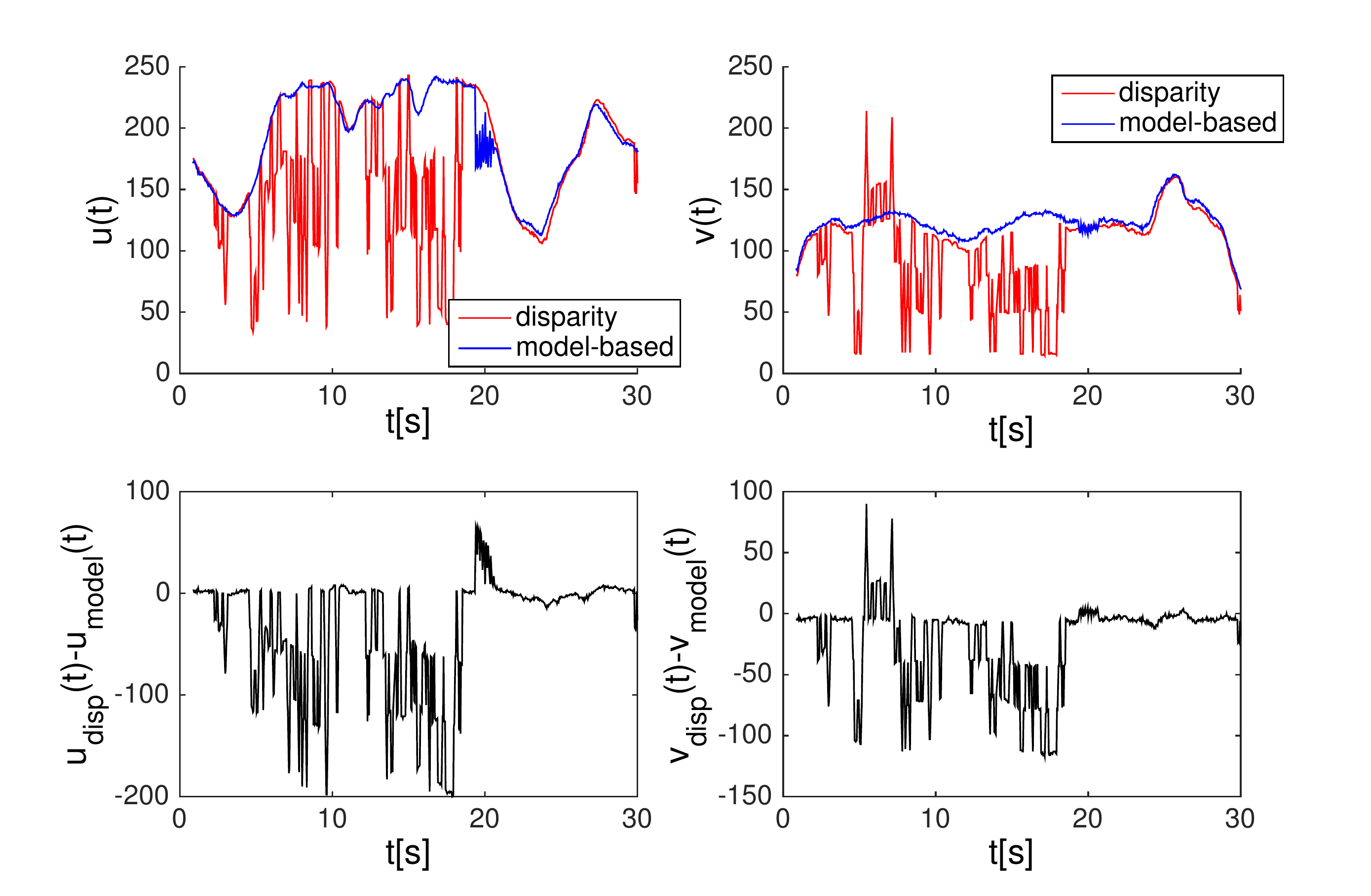}
\caption{The same as Fig.~\ref{fig:dataVSblobs_acquired}, using SGBM instead of LIBELAS. The blue trace is the same, the red trace is obtained offline, computing disparity on the same acquired sequence of Fig.~\ref{fig:dataVSblobs_acquired}, in order to compare the two on the same frames.}
\label{fig:dataVSblobs_offline_SGBM}
\end{figure}

Fig.~\ref{fig:dataVSblobs_offline_SGBM} shows the same quantities of Fig.~\ref{fig:dataVSblobs_acquired} but computed on the disparity map provided by SGBM. We computed the disparity map offline on the same set of images acquired when tracking with the ELAS algorithm. As can be clearly noticed, the unstable behavior of the disparity produced by SGBM is not sufficient to provide a fast and reliable signal to track the ball.


\section{Applications of Disparity on Humanoid Robots}
\label{sec:applications}

In this section we show how the depth-driven attention system described in Sec~\ref{sec:results} can be employed to improve the robot perception of the surrounding environment. In particular we consider a basic interaction between the robot and a human teacher or the situation where the robot needs to visually parse the objects lying on a table. Our observations are mainly qualitative in this section.

\subsection{On the fly object recognition}

One of the most natural applications of the presented system is the extension to an interaction framework previously proposed in the context of (visual) object learning and recognition~\cite{fanello13b}. We consider a setting where a human teacher shows novel objects to the iCub in order for the robot to focus its attention towards them and therefore learn their visual appearance. Communication between the human and the robot occurs through speech, i.e. commands and object labels are verbally provided by the human teacher (see \cite{fanello13c} for more details about a thorough overview of the system).

\begin{figure}[t]
\centering
\includegraphics[width=1.0\linewidth]{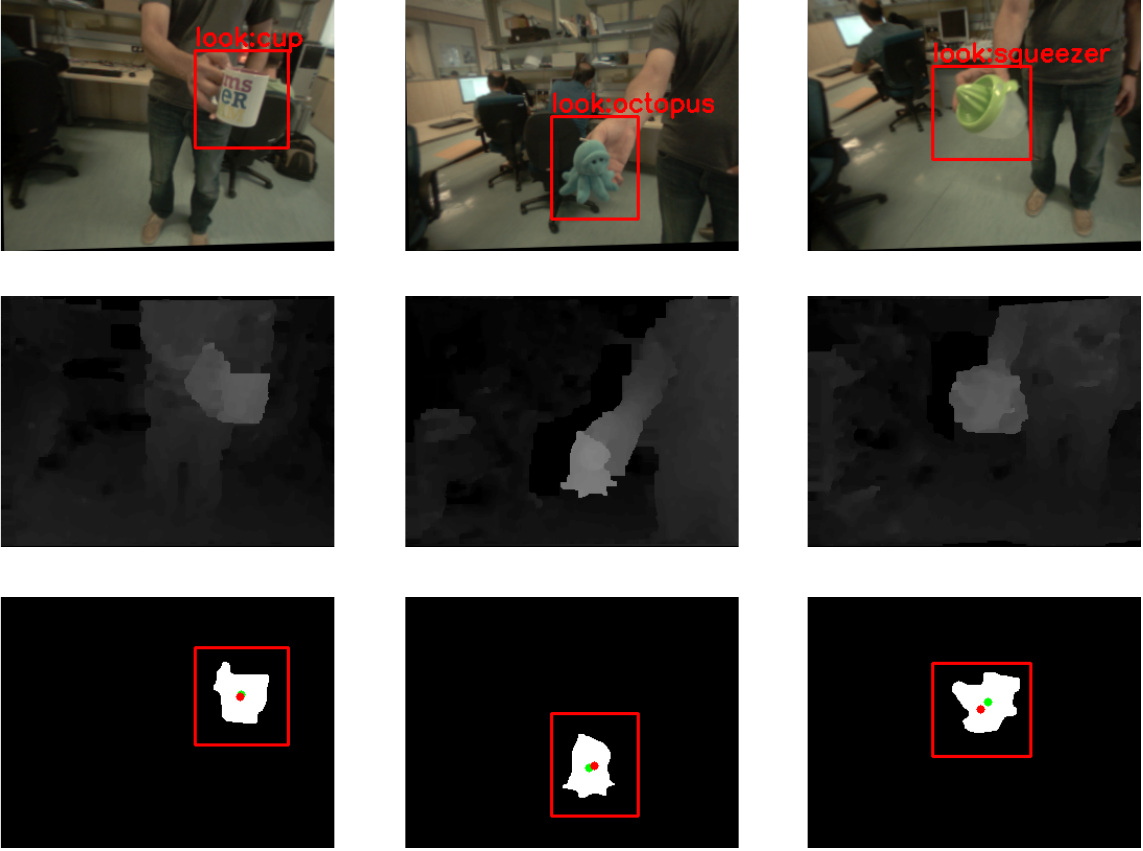}
\caption{Three frames (extracted from the attached video) showing the effectiveness of the proposed segmentation system. Top: resulting crop in the left rectified frame, labeled by the operator's verbal supervision. Middle: disparity map. Bottom: segmented disparity blob, its centroid and the enclosing ROI.}
\label{fig:onthefly}
\end{figure}

In Fig.~\ref{fig:onthefly} we report three frames extracted from three corresponding sequences, recorded while tracking three different objects following the disparity-based strategy previously described. The top row shows the output of the pipeline: the object (in this case a cup, a toy octopus and a lemon squeezer) is localized in the scene using the ROI provided by the disparity segmentation module, the label being provided verbally by the human teacher. The middle row reports the associated disparity map by ELAS and the bottom row reports its segmentation, together with the centroid, that is used for the tracking (red dot: average over three frames, green dot: current frame) and the ROI, used for the segmentation (averaged over three frames to account for spurious mis-segmentations). The ROI is computed as the smallest rectangular region enclosing the segmented blob, with a margin of $20$ pixels. The video acquired from the iCub cameras during this experiment is attached to the paper. Observing the stability exhibited by the depth-driven attention strategy, we can conclude that this is clearly a viable alternative to the motion-based tracker employed in~\cite{fanello13b,fanello13c,ciliberto13}. Moreover, it can be observed that, since the disparity cue does not require the human teacher to continuously shake the object of interest in front of the robot as when using motion information, now the object can be kept still or moved more slowly. This results in a more accurate segmentation, being also possible to average over a time window. Indeed, we are currently collecting a large-scale visual recognition dataset for robotics with this same application. 

\subsection{Object exploration and manipulation}

Finally we consider a setting in which the robot is standing in front of a table and uses disparity to distinguish separate objects in order then to perform more complex actions such as, learn their appearance, reach for them and eventually grasp them. In particular, using the disparity map we could reconstruct the scene in front of the robot and the system could determine the optimal hand pose for a reliable grasp~\cite{gori2014}. 

\begin{figure}[t]
\centering
\includegraphics[width=1.0\linewidth]{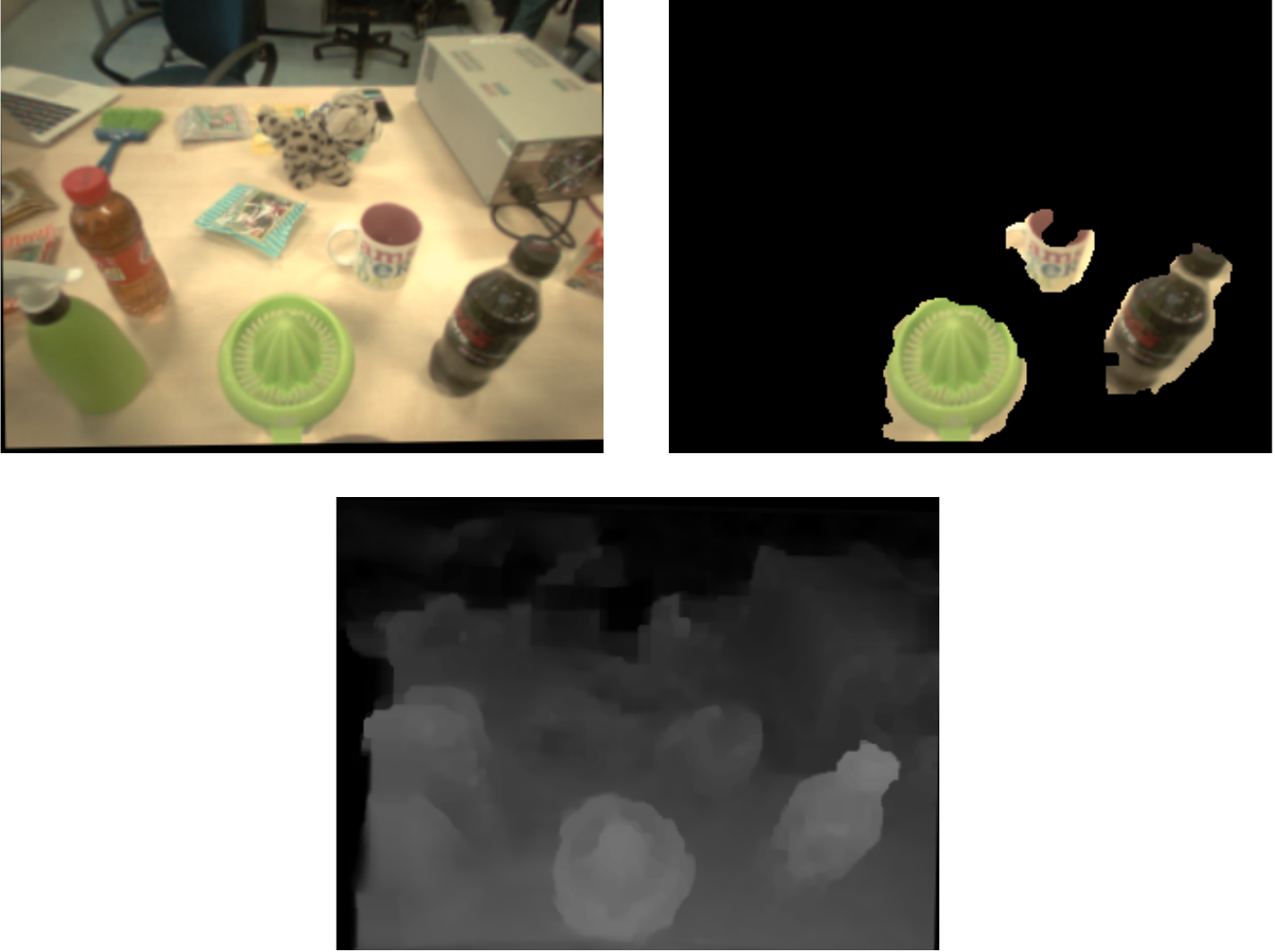}
\caption{Top left: rectified frame recorded from the iCub's left camera while the robot was looking at a table in front of it. Top right: segmentation of the three closest objects on the table obtained from the disparity map of the scene (bottom).}
\label{fig:iol}
\end{figure}

In Fig.~\ref{fig:iol}, we report the left rectified image (top left) and the corresponding segmentation (top right), obtained by putting a threshold on the disparity map (bottom). For the purpose of demonstration such threshold was chosen manually, however in a real application more sophisticated processing of the disparity map could applied to cluster $3$D point clouds and better detect separate objects.

\section{Conclusions}

In this work we have described the current system implemented on the iCub robot to perform depth estimation and how it benefits from the recent incorporation of the state-of-the-art disparity computation algorithm ELAS~\cite{LIBELAS}. We have evaluated some real applications of the information provided by the disparity map produced by ELAS to typical robotics settings, pointing out that this approach is indeed computationally efficient and robust for the real-world scenario. We have therefore observed that depth information could now be used at the basis of more complex behaviors of the humanoid robotic system, such as interaction with the human or with the surrounding environment. The system described in this paper for depth estimation is already available for the iCub community and can be used as an off-the-shelf solution for all iCub users. Soon also the application for disparity-driven attention will be made publicly available.

\addtolength{\textheight}{-11cm}   





\bibliographystyle{IEEEtranBST/IEEEtran}
\bibliography{IEEEtranBST/IEEEabrv,./biblio}

\end{document}